\documentclass[conference]{IEEEtran}

\usepackage{amsmath,amssymb}
\usepackage{algorithmic}
\usepackage{algorithm}
\usepackage{siunitx}
\usepackage{bm}
\newcolumntype{C}{>{\centering\arraybackslash}X}
\usepackage{booktabs}
\usepackage{tabularx}
\usepackage{graphicx}
\usepackage{hyperref}
\usepackage{xcolor}

\newcommand{\R}{\mathbb{R}}
\newcommand{\vect}[1]{\mathbf{#1}}
\DeclareMathOperator*{\softmax}{softmax}
\DeclareMathOperator*{\sigmoid}{sigmoid}

\newcommand{\firstpage}[1]{}
\newcommand{\pubvolume}[1]{}
\newcommand{\issuenum}[1]{}
\newcommand{\articlenumber}[1]{}
\newcommand{\pubyear}[1]{}
\newcommand{\copyrightyear}[1]{}
\newcommand{\doinum}[1]{}
\newcommand{\datereceived}[1]{}
\newcommand{\daterevised}[1]{}
\newcommand{\dateaccepted}[1]{}
\newcommand{\datepublished}[1]{}
\newcommand{\AuthorNames}[1]{}
\newcommand{\keyword}[1]{\noindent\textbf{Keywords:} #1}
\newcommand{\authorcontributions}[1]{}
\newcommand{\funding}[1]{\section*{Funding}#1}
\newcommand{\institutionalreview}[1]{\section*{Institutional Review Board Statement}#1}
\newcommand{\informedconsent}[1]{\section*{Informed Consent Statement}#1}
\newcommand{\dataavailability}[1]{\section*{Data Availability Statement}#1}
\newcommand{\acknowledgments}[1]{\section*{Acknowledgments}#1}
\newcommand{\conflictsofinterest}[1]{\section*{Conflicts of Interest}#1}
\newcommand{\abbreviations}[2]{\section*{#1}#2}
\newcommand{\reftitle}[1]{}

\title{Regime-Aware Physics-Guided Early Warning of Lithium-Ion Battery Thermal Runaway Using Thermo-Mechanical Signals}

\author{Syed Sajid Ullah\thanks{Correspondence: sajid@chd.edu.cn},
Muhammad Zunair Zamir and Salman Khan}

\begin{document}

\maketitle

\begin{abstract}
Thermal runaway in lithium-ion batteries poses a major safety risk to electric vehicles and energy storage systems.  Current early-warning methods depend mainly on temperature and may therefore miss mechanical precursors that emerge before rapid heating.  We introduce a regime-aware, physics-guided framework that integrates temperature, voltage, force, deformation, and state-of-charge measurements for early warning under controlled mechanical abuse.  A lightweight convolutional classifier first infers safe, warning, or danger regimes from mechanical signals.  These regime estimates then condition a causal temporal convolutional backbone through feature-wise linear modulation, physics-biased attention, and regime-dependent gating.  Joint learning unifies regime identification, thermal-runaway detection, and time-to-disaster estimation.  We evaluate the framework using leave-one-experiment-out cross-validation on 30 mechanical-abuse tests across state-of-charge levels of 10\%, 50\%, and 90\% and two loading protocols.  The method achieves an F1 score of 0.89, a high-temperature prediction root-mean-square error of \SI{12.3}{\degreeCelsius}, a mean warning lead time of \SI{15.6}{s}, a detection success rate of 0.92, and an experiment-level false alarm rate of 2.7\%.  Its lead time exceeds that of the strongest baseline by 69.6\%.  Removing force reduces the lead time by 60.3\%, highlighting the value of mechanical precursors.  These results support regime-aware thermo-mechanical fusion as a promising strategy for earlier and more reliable thermal-runaway warning under controlled abuse conditions.
\end{abstract}

\keyword{battery thermal runaway; early warning; physics-guided neural networks; temporal convolutional network; feature-wise linear modulation; multi-task learning; leave-one-experiment-out cross-validation}

\section{Introduction}

Lithium-ion batteries (LIBs) are now central to electric vehicles, grid storage, and portable electronics because of their high energy density and long cycle life.  However, LIBs continue to face severe safety challenges associated with thermal instability, internal short circuits, overheating, and thermal runaway (TR), which remain among the most catastrophic failure mechanisms in modern battery systems~\cite{kong2023review,xu2021mitigation}.

Thermal runaway refers to a highly nonlinear electrochemical-thermal process in which self-generated heat exceeds the dissipation capability of the battery system, resulting in uncontrollable temperature escalation, gas venting, fire propagation, toxic emissions, and explosions.  The initiation of thermal runaway can occur because of overcharging, over-discharging, mechanical deformation, separator failure, lithium dendrite growth, manufacturing defects, or external thermal abuse~\cite{liu2024effect}.  As battery energy density increases, the thermal tolerance margin becomes progressively smaller, thereby intensifying safety risks and increasing the complexity of battery-management systems (BMSs).

To address these challenges, researchers have extensively investigated thermal-runaway early-warning methods using thermal, electrical, electrochemical, mechanical, acoustic, and gas-emission signals.  Among these approaches, gas-sensing-based monitoring has shown strong potential because gaseous emissions typically occur prior to measurable voltage collapse or abnormal temperature rise.  Wang et al.~\cite{wang2024monitoring} investigated gas-sensor-based thermal-runaway monitoring using CO\textsubscript{2}, H\textsubscript{2}, CO, and volatile organic compounds, demonstrating that gas signatures can provide effective early indicators of abnormal electrochemical decomposition.  Similarly, Cui et al.~\cite{cui2023thermal} proposed a gas-production-based thermal-runaway early-warning framework and introduced a thermal-runaway severity index capable of estimating the evolution and risk level of battery failure before catastrophic runaway occurs.

In addition to gas-emission analysis, electrochemical and impedance-based monitoring approaches have gained significant attention for battery fault diagnosis.  Dong et al.~\cite{dong2021reliable} demonstrated that electrochemical impedance spectroscopy (EIS) can provide reliable precursor information regarding internal degradation and thermal instability.  Lyu et al.~\cite{lyu2021real} further proposed an online impedance-measurement strategy for overcharge warning and thermal-runaway prediction, showing that impedance variations can serve as early indicators of abnormal electrochemical reactions.

Online data-driven fault-diagnosis frameworks have also demonstrated strong capability for thermal-runaway prediction.  Sun et al.~\cite{sun2022online} proposed an online thermal-runaway early-warning framework using operational electrical signals from EV batteries and showed that intelligent monitoring can significantly improve fault-detection accuracy under dynamic operating conditions.  Gao et al.~\cite{gao2020case} analyzed practical EV battery thermal-runaway cases and emphasized the importance of online internal-short-circuit detection for preventing catastrophic failures in battery packs.

The rapid development of machine learning and deep learning has further transformed battery fault diagnosis and prognostics.  Khaleghi et al.~\cite{khaleghi2022developing} proposed a data-driven prognostics and health-management framework for lithium-ion batteries and highlighted the importance of intelligent data analytics in modern BMSs.  More recently, Chen et al.~\cite{chen2025model} developed a model-constrained deep-learning framework capable of robust online fault diagnosis under stochastic operating conditions.  Their study demonstrated that incorporating physical battery constraints into deep neural networks substantially improves generalization and interpretability.

Recent studies have also demonstrated that forecasting-oriented architectures can substantially improve battery anomaly prediction and long-horizon temporal representation learning.  Zeng et al.~\cite{zeng2023transformers} argued that simple linear forecasting architectures can outperform complex transformer models in several long-sequence forecasting tasks, highlighting the importance of efficient inductive bias design for temporal learning systems.  In addition, physics-based learning strategies have shown strong capability for safety-critical battery applications.  Firoozi et al.~\cite{firoozi2021cylindrical} proposed a physics-based learning framework for cylindrical battery fault detection under extreme fast-charging conditions and demonstrated improved robustness for abnormal battery-state identification.  Such studies indicate that combining physical battery constraints with modern forecasting architectures can significantly improve reliability and interpretability in thermal-runaway early-warning systems.

Transfer-learning-based diagnosis frameworks have also emerged as promising solutions for limited-data environments.  Dong and Sun~\cite{dong2024multi} proposed a multi-source domain-transfer-learning framework for thermal-runaway diagnosis, enabling knowledge transfer across different battery chemistries and operational conditions.  Such approaches are particularly important because large-scale thermal-runaway experiments are expensive, hazardous, and difficult to conduct repeatedly.

Besides fault diagnosis, thermal-runaway mitigation and propagation suppression have become important research directions.  Huang et al.~\cite{huang2025mechanism} investigated aerogel-based insulation materials for suppressing thermal propagation in lithium-ion battery modules and established a safety-evaluation methodology for high-energy-density systems.  Zhang et al.~\cite{zhang2022non} proposed a non-uniform phase-change-material strategy for directional mitigation of thermal-runaway propagation and demonstrated significant improvements in thermal isolation capability.  These studies indicate that thermal management and propagation suppression must complement early-warning systems to achieve comprehensive battery safety.

Several researchers have also investigated probabilistic risk-analysis frameworks for battery safety assessment.  Meng et al.~\cite{meng2023integrated} proposed an integrated methodology for dynamic thermal-runaway risk prediction using Bayesian networks and support-vector regression.  Their later work incorporated physics-informed Bayesian networks for uncertainty-aware battery accident analysis~\cite{meng2024risk}.  Similarly, Wang et al.~\cite{wang2026real} proposed a Bayesian fault-propagation framework for real-time reliability analysis of battery-energy-storage systems, providing an interpretable mechanism for safety-risk evolution modeling.

The emergence of deep-learning-based forecasting models has opened new opportunities for intelligent battery-safety prediction.  Rather than surveying every recent forecasting architecture, the present work focuses on architectures that are practical for short-horizon, causal, multivariate warning under mechanical abuse.  This scope motivates the use of recurrent, convolutional, transformer, and TCN baselines in the experiments.

Despite substantial progress in battery fault diagnosis and temporal forecasting, several challenges remain unresolved.  Existing thermal-runaway prediction frameworks often suffer from limited generalization capability, insufficient interpretability, weak robustness under varying operational conditions, and inadequate integration between physics-based battery dynamics and deep-learning architectures.  Moreover, many existing methods rely primarily on electrical measurements while ignoring thermo-mechanical interactions and structural degradation mechanisms that may provide valuable precursor information regarding fault evolution.

Therefore, there remains a strong need for regime-aware and physics-guided early-warning frameworks capable of integrating thermo-mechanical sensing, multivariate temporal forecasting, and intelligent fault diagnosis for robust TR prediction.  Motivated by these challenges, this work proposes a physics-guided deep-learning framework for LIB TR early warning using thermo-mechanical signals and causal temporal modeling.  The proposed framework aims to improve prediction lead time, interpretability, robustness, and operational reliability under dynamic battery conditions.

This manuscript is distinct from the authors' companion study on gradient-boosting-based infrared-hotspot warning.  That study used image-derived hotspot descriptors and tree-based decision rules as the primary modeling route, whereas the present paper uses synchronized temperature, force, deformation, voltage, and SOC information in a regime-aware neural architecture with multitask TR detection and time-to-disaster estimation.  The overlap is therefore limited to the broader mechanical-abuse safety context and, where applicable, shared experimental infrastructure; the modeling objective, input representation, and claimed contribution are different.

The main contributions of this work are summarized as follows:
\begin{itemize}
  \item Thermal runaway early warning under mechanical abuse is formulated as a causal thermo-electro-mechanical sequence prediction problem, providing a principled foundation for precursor-driven failure forecasting.
  \item A regime-aware temporal modeling framework is proposed that leverages mechanical precursor signals to identify transitions among safety states, enabling adaptive warning thresholds aligned with the battery's physical condition.
  \item A physics-guided, SOC-conditioned feature modulation strategy is introduced for joint warning-state classification and time-to-danger estimation, incorporating electrochemical priors through learnable FiLM-based conditioning.
  \item An experiment-level evaluation protocol is established using 30 real mechanical abuse experiments with leave-one-experiment-out validation and early-warning metrics, including lead time, false-alarm rate, missed-alarm rate, and detection success rate.
\end{itemize}

The remainder of this paper is organized as follows.  Section~\ref{sec:related} reviews TR warning methods, mechanical abuse studies, and physics-guided learning approaches for battery safety.  Section~\ref{sec:formulation} defines the causal early-warning problem, TR onset criterion, regime labels, and prediction targets.  Section~\ref{sec:method} presents the proposed regime-aware physics-guided temporal learning framework.  Section~\ref{sec:experiments} describes the dataset, preprocessing, baselines, and evaluation protocol.  Section~\ref{sec:results} reports the quantitative results, ablation analysis, and interpretability study.  Finally, Section~\ref{sec:conclusion} concludes the paper and discusses future research directions.

\section{Related Work}
\label{sec:related}

\subsection{Thermal Runaway Early Warning and Fault Diagnosis}

Thermal-runaway early warning has attracted significant research attention because conventional battery-management systems are often unable to provide sufficiently early detection of catastrophic battery failures.  Existing thermal-runaway diagnosis approaches can generally be classified into gas-based monitoring, impedance-based methods, probabilistic risk-analysis frameworks, deep-learning-based fault diagnosis, and hybrid physics-guided prediction systems.

Gas-sensing mechanisms are among the earliest indicators of thermal instability because gaseous byproducts are typically released before severe temperature escalation occurs.  Wang et al.~\cite{wang2024monitoring} investigated gas-emission monitoring using chemical sensors and demonstrated that gases such as CO\textsubscript{2}, H\textsubscript{2}, and CO can effectively indicate abnormal electrochemical decomposition during early-stage thermal runaway.  Cui et al.~\cite{cui2023thermal} extended this concept by introducing a thermal-runaway severity index based on gas-production characteristics and demonstrated that gas evolution can provide significantly earlier warning compared with voltage or temperature measurements.  In addition, Tam et al.~\cite{tam2024development} developed an acoustic-based early-stage thermal-runaway detection framework using machine-learning analysis and demonstrated that abnormal acoustic signatures can serve as precursor indicators before catastrophic battery failure occurs.

Electrochemical impedance spectroscopy has also shown strong potential for early fault diagnosis because impedance variations reflect internal electrochemical degradation processes.  Dong et al.~\cite{dong2021reliable} proposed an EIS-based thermal-runaway warning framework and demonstrated reliable detection of abnormal electrochemical behavior.  Similarly, Lyu et al.~\cite{lyu2021real} developed an online impedance-measurement strategy for overcharge warning and thermal-runaway prediction under dynamic charging conditions.  These studies indicate that electrochemical signatures can provide highly sensitive information regarding early-stage battery degradation.

Electrical-signal-based monitoring remains one of the most practical approaches for real-time battery fault diagnosis.  Sun et al.~\cite{sun2022online} proposed an online data-driven thermal-runaway warning framework using operational EV battery signals and demonstrated accurate fault detection during dynamic operating conditions.  Gao et al.~\cite{gao2020case} investigated practical EV battery thermal-runaway cases and analyzed online internal-short-circuit detection methods for preventing catastrophic failure propagation.

Several recent studies have explored hybrid deep-learning architectures for thermal-runaway prediction under dynamic charging and operational conditions.  Huang et al.~\cite{huang2025study} proposed a convolutional-transformer-based early-warning framework capable of modeling nonlinear thermal dynamics during high-rate charging/discharging scenarios.  Their framework demonstrated strong predictive capability under rapidly varying thermal conditions.  Similarly, Wang et al.~\cite{wang2025fault} employed convolutional neural networks for onboard thermal-runaway fault-cause analysis and demonstrated improved post-failure diagnosis capability for lithium-ion battery systems.

Deep-learning-based battery diagnosis frameworks have shown remarkable progress in recent years.  Chen et al.~\cite{chen2025model} proposed a model-constrained deep-learning framework for online lithium-ion battery fault diagnosis under stochastic conditions.  Their method integrated physical battery constraints into neural-network training and achieved substantial improvements in robustness and interpretability.  Fan et al.~\cite{fan2025fault} proposed a feature-augmented attentional autoencoder for battery fault detection and demonstrated that attention mechanisms can effectively capture abnormal temporal behavior in multivariate battery signals.

Graph-neural-network-based approaches have also gained considerable attention in battery-pack diagnostics.  Ouyang et al.~\cite{ouyang2025voltage} developed an optimized graphical neural network for voltage-fault diagnosis in EV batteries and demonstrated improved fault-localization capability compared with conventional deep-learning methods.  These approaches highlight the importance of structural and relational modeling in large-scale battery systems.

Transfer learning and domain adaptation techniques have emerged as promising solutions for addressing limited-data challenges in battery-fault diagnosis.  Dong and Sun~\cite{dong2024multi} proposed a multi-source domain-transfer-learning framework with few-shot learning capability for thermal-runaway diagnosis across different battery chemistries and operating environments.  Their framework significantly improved diagnosis generalization under limited labeled data conditions.

Battery-state prediction and health-estimation studies have also contributed significantly to thermal-runaway prevention because battery degradation strongly influences thermal stability and internal short-circuit probability.  Wang et al.~\cite{wang2024imfo} proposed an IMFO-LSTM-BiGRU framework for long-term multi-state battery prediction and demonstrated improved forecasting accuracy under varying operating conditions.  Furthermore, Wang et al.~\cite{wang2025deep} introduced a deep-learning framework for battery state-of-health estimation using partial charging data and polarization-equilibrium analysis, achieving high estimation accuracy with low computational overhead.  Khaleghi et al.~\cite{khaleghi2022developing} further emphasized the importance of intelligent prognostics and health-management frameworks for modern lithium-ion battery systems.

Several studies have also focused on thermal-runaway mitigation and propagation suppression.  Huang et al.~\cite{huang2025mechanism} investigated aerogel-based insulation materials for suppressing thermal propagation in battery modules, while Zhang et al.~\cite{zhang2022non} proposed a non-uniform phase-change-material strategy for directional thermal mitigation.  Liu et al.~\cite{liu2024effect} analyzed the influence of external thermal conditions on thermal-runaway initiation and propagation in cylindrical lithium-ion batteries.  These studies collectively demonstrate that thermal management and propagation suppression remain critical components of battery-safety design.

Probabilistic risk-analysis frameworks provide additional interpretability for battery safety analysis.  Meng et al.~\cite{meng2023integrated} proposed a dynamic risk-prediction methodology based on Bayesian networks and support-vector regression, while their later work introduced physics-informed Bayesian networks for uncertainty-aware battery accident analysis~\cite{meng2024risk}.  Wang et al.~\cite{wang2026real} further developed a Bayesian fault-propagation framework for real-time reliability analysis of battery-energy-storage systems.

\subsection{Temporal Learning and Physics-Guided Modeling}

Modern sequence-learning methods provide useful tools for battery early warning, but a TR warning model must remain causal, short-horizon, and reliable under limited destructive-test data.  For this reason, we compare against compact LSTM, CNN-LSTM, Transformer, and TCN baselines rather than importing large long-horizon forecasting architectures whose assumptions and data requirements differ from the present mechanical-abuse setting.  Recent work on temporal inductive bias and physics-based learning supports this design choice: Zeng et al.~\cite{zeng2023transformers} showed that simpler temporal models can outperform complex transformers in some forecasting settings, while Firoozi et al.~\cite{firoozi2021cylindrical} demonstrated the value of physics-based learning for battery fault detection under extreme conditions.

Table~\ref{tab:related_work} compares representative related studies on thermal-runaway early warning.  Several key gaps emerge from the table.  Existing methods predominantly rely on temperature, voltage, or gas signals with no integration of mechanical precursors such as force or deformation.  They also lack regime-aware processing to distinguish safe, warning, and danger states.  Furthermore, these approaches either introduce additional hardware overhead or suffer from high computational complexity.  To address these limitations, this paper proposes a regime-aware, physics-guided deep learning framework that fuses thermo-mechanical signals including temperature, force, deformation, and voltage for early thermal-runaway warning.  The proposed method is detailed in the following sections.

\begin{table}[t]
\centering
\caption{Comparison of representative related studies on lithium-ion battery thermal-runaway early warning.}
\label{tab:related_work}
\footnotesize
\resizebox{\columnwidth}{!}{
\begin{tabularx}{\columnwidth}{c c p{2cm} p{3.5cm}}
\toprule
Year & Ref. & Method & Limitations \\
\midrule
2022 & {[}8{]} & Online data-driven fault diagnosis using voltage, current, and temperature. & Relies on electrical and thermal signals; limited interpretability. \\
2023 & {[}5{]} & Gas-production-based TR early-warning and severity estimation. & Requires additional gas-sensing hardware. \\
2023 & {[}17{]} & Dynamic risk prediction with fault-tree analysis and Bayesian networks. & Limited real-time deep-learning capability. \\
2024 & {[}4{]} & Gas-sensor-based TR monitoring using CO\textsubscript{2}, H\textsubscript{2}, CO. & Increases system cost; requires robust calibration. \\
2024 & {[}14{]} & Multi-source domain-transfer learning for TR diagnosis. & May suffer from negative transfer under domain mismatch. \\
2025 & {[}11{]} & Model-constrained deep learning for online fault diagnosis. & Model design and training more complex than data-driven methods. \\
2025 & {[}33{]} & Feature-augmented attentional autoencoder for battery fault detection. & Detection sensitive to training-data quality. \\
2025 & {[}31{]} & Convolutional-transformer TR early-warning for high-rate conditions. & High computational cost for embedded BMS deployment. \\
2025 & {[}34{]} & Optimized GNN for voltage-fault diagnosis in EV battery packs. & Requires pack-topology information. \\
2026 & {[}19{]} & Bayesian fault-propagation network for real-time reliability analysis. & Probabilistic inference computationally demanding. \\
\bottomrule
\end{tabularx}
}
\end{table}

\section{Problem Formulation}
\label{sec:formulation}

We define the thermal runaway onset time $t_{\mathrm{TR}}$ as the earliest time at which either of two conditions is satisfied:
\begin{equation}
  \label{eq:tr_onset}
  t_{\mathrm{TR}} = \min\bigl\{t : T(t) \ge T_{\mathrm{TR}} \;\lor\; \dot{T}(t) \ge \dot{T}_{\mathrm{TR}}\bigr\}
\end{equation}
where $T(t)$ denotes the cell surface temperature, $\dot{T}(t) = dT/dt$ the temperature rate of rise, $T_{\mathrm{TR}} = \SI{150}{\degreeCelsius}$ the absolute temperature threshold, and $\dot{T}_{\mathrm{TR}} = \SI{3}{\degreeCelsius/s}$ the rate-of-rise threshold.  These dual criteria capture both gradual temperature escalation (to \SI{150}{\degreeCelsius}) and rapid exothermic acceleration (exceeding \SI{3}{\degreeCelsius/s}), which are the two canonical signatures of TR onset established in the battery safety literature~\cite{feng2018thermal}.

\begin{figure}[t]
  \centering
  \includegraphics[width=8.5cm]{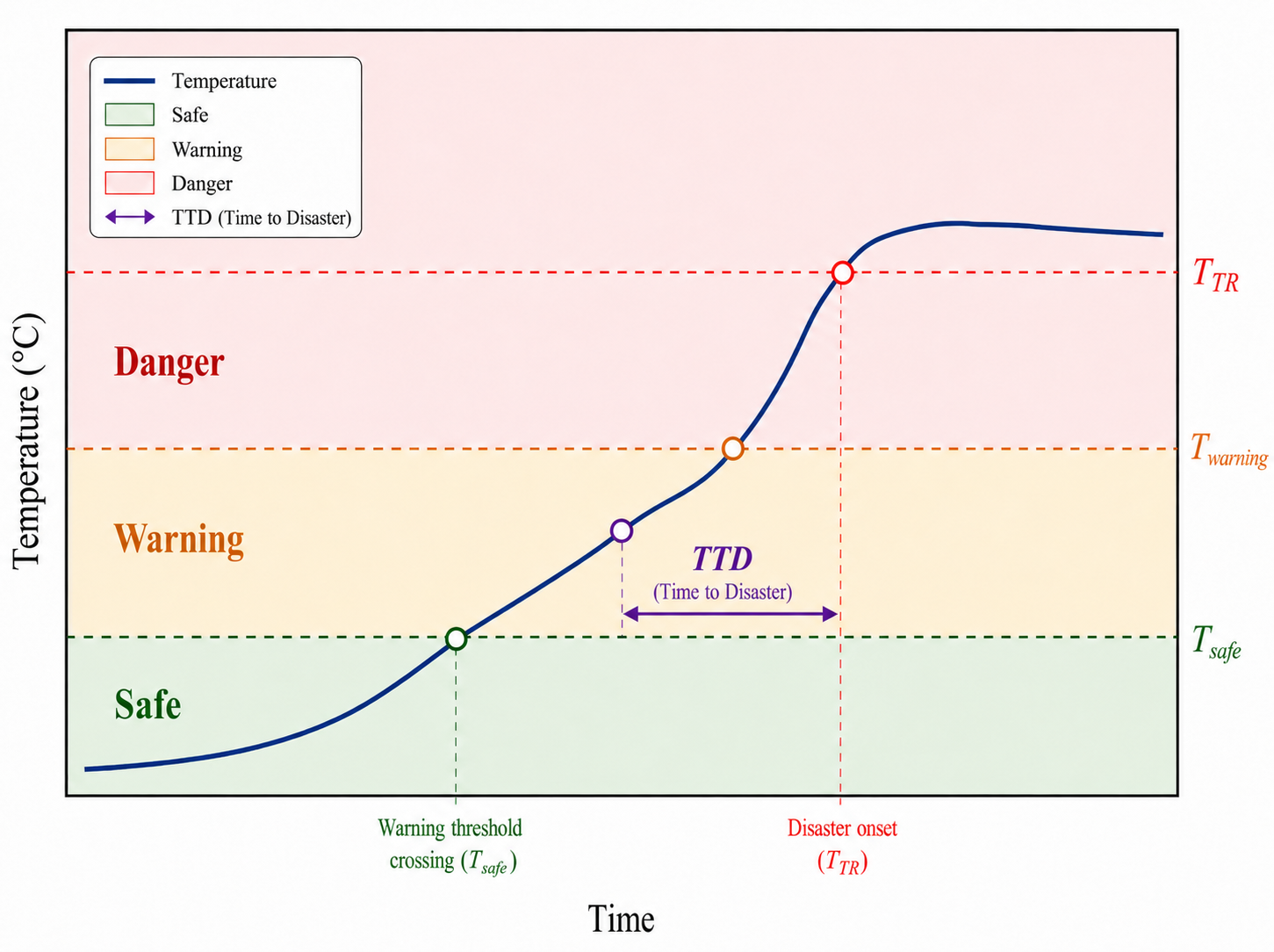}
  \caption{Timeline of thermal progression through the safety regimes to TR onset.  $L_{\mathrm{lead}}$ denotes the interval between the model's first valid warning and the measured TR onset.}
  \label{fig:problem_timeline}
\end{figure}

The battery's thermal trajectory is partitioned into three mutually exclusive safety regimes based on temperature thresholds, as illustrated in Fig.~\ref{fig:problem_timeline}:
\begin{equation}
  \label{eq:regime}
  r(t) =
  \begin{cases}
    0 \;\text{(Safe)},    & T(t) < T_{\mathrm{safe}} \\
    1 \;\text{(Warning)}, & T_{\mathrm{safe}} \le T(t) < T_{\mathrm{danger}} \\
    2 \;\text{(Danger)},  & T(t) \ge T_{\mathrm{danger}}
  \end{cases}
\end{equation}
where $T_{\mathrm{safe}} = \SI{60}{\degreeCelsius}$ and $T_{\mathrm{danger}} = \SI{120}{\degreeCelsius}$.  These fixed thresholds are applied identically across SOC levels and loading protocols.  The safe regime corresponds to normal operating temperatures where no safety concern exists.  The warning regime captures the early onset of anomalous heating, during which internal exothermic reactions may have begun but remain controllable through active cooling.  The danger regime indicates that the cell is on an irreversible trajectory toward TR and immediate protective action is required.

The regime labels are generated algorithmically from synchronized temperature measurements after timestamp alignment and smoothing; no fold-specific or experiment-specific threshold tuning is used.  Force and deformation are not used to define the ground-truth regime labels, which prevents leakage from the mechanical channels into the target definition.  They are used only as model inputs, allowing the Stage~1 classifier to learn whether mechanical precursor patterns anticipate the temperature-defined safety state.

The lead time $L_{\mathrm{lead}}$ quantifies how far in advance the model can issue a TR warning relative to the actual TR onset:
\begin{equation}
  \label{eq:lead_time}
  L_{\mathrm{lead}} = t_{\mathrm{TR}} - t_{\mathrm{warn}}
\end{equation}
where $t_{\mathrm{warn}}$ is the time at which the model first correctly predicts an imminent TR event (i.e., TR probability exceeds a detection threshold of 0.5).  A larger lead time provides more opportunity for mitigative actions, making it the most operationally significant metric for TR early-warning systems.

Given a multivariate time-series window $\mathbf{X} \in \R^{W \times C}$ of length $W$ with five synchronized sensor channels (high temperature, low temperature, voltage, force, and deformation) and a scalar state-of-charge $s \in [0,1]$, the model produces three outputs:
\begin{enumerate}
  \item Regime classification: $\hat{\vect{r}} = f_r(\mathbf{X}, s) \in \R^{3}$, predicting the probability distribution over safety regimes at the current time step.
  \item TR detection: $\hat{y}_{\mathrm{TR}} = f_d(\mathbf{X}, s) \in [0,1]$, estimating the probability that TR will occur within the prediction horizon of $H = 60$ s.
  \item Time-to-disaster regression: $\hat{t}_{\mathrm{TTD}} = f_t(\mathbf{X}, s) \in \R_{\ge 0}$, predicting the remaining time (in seconds) until TR onset.
\end{enumerate}

An effective TR early-warning system must satisfy three operational requirements:
\begin{itemize}
  \item Sufficient lead time: $L_{\mathrm{lead}} > \SI{10}{s}$ to enable protective actions.
  \item High detection rate: Detection success rate $\mathrm{DSR} > 85\%$.
  \item Low false alarm rate: experiment-level $\mathrm{FAR} < 10\%$ of held-out test experiments.
\end{itemize}

\section{Proposed Method}
\label{sec:method}

\subsection{System Overview}

\begin{figure*}[t]
\centering
\includegraphics[width=\textwidth]{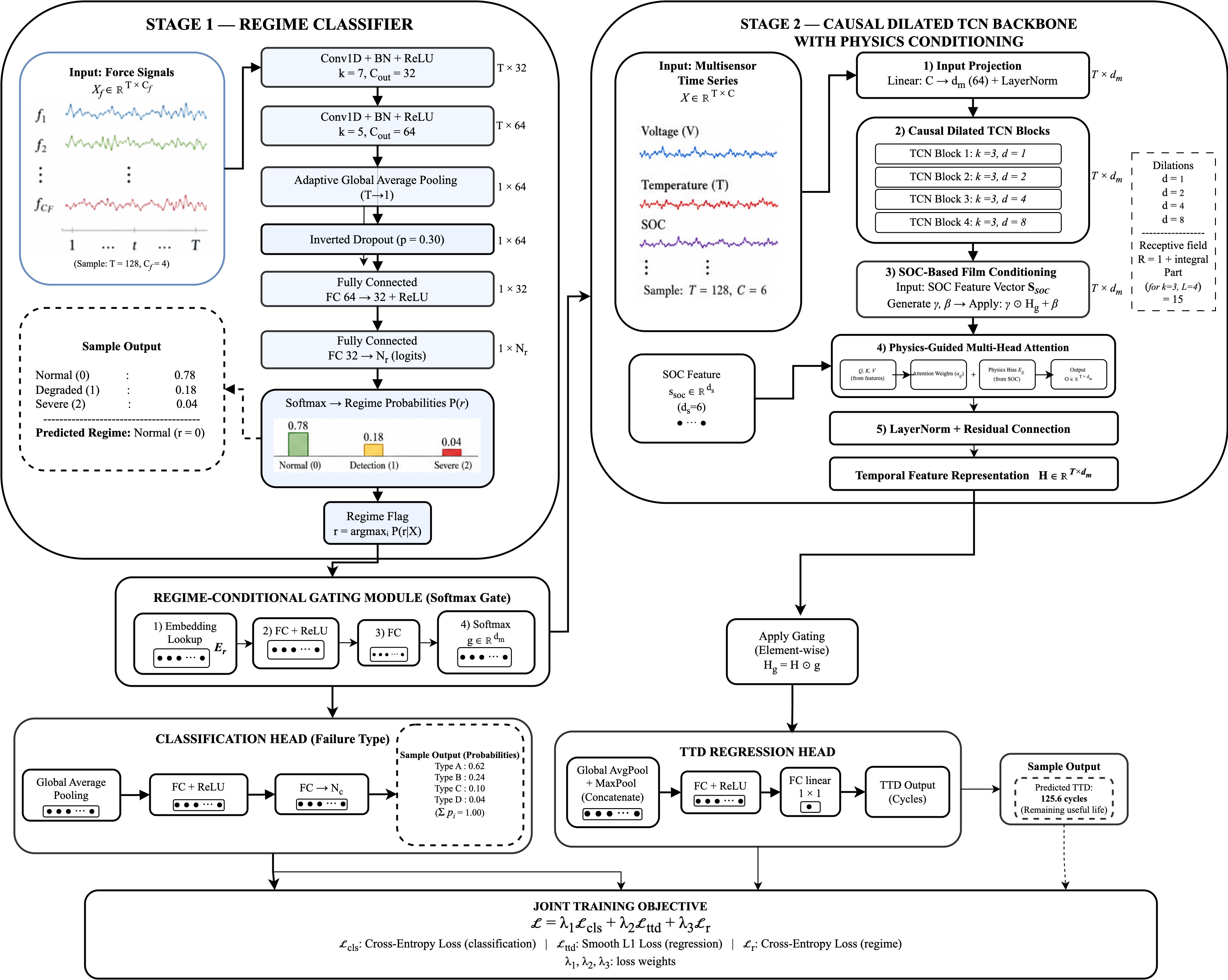}
\caption{Overall architecture of the proposed regime-aware, physics-guided TR early-warning framework. Stage~1 uses a lightweight CNN to estimate the operating regime from force signals. Stage~2 combines a causal dilated TCN backbone with SOC-based FiLM conditioning, physics-guided attention, and regime-conditional gating, followed by classification and time-to-disaster regression heads trained with a joint objective.}
\label{fig:architecture}
\end{figure*}

Fig.~\ref{fig:architecture} presents the overall architecture of the proposed regime-aware, physics-guided TR early-warning framework.  The system comprises two stages operating in cascade.  Stage~1 is a lightweight CNN-based regime classifier that processes the force and deformation channels to produce a probability distribution over the three safety regimes.  Stage~2 is a physics-guided temporal backbone that ingests the full five-channel multivariate input and produces three task-specific outputs---regime classification logits, TR detection probability, and TTD regression value---while being conditioned on both the SOC (through FiLM) and the regime probabilities from Stage~1 (through gating).

\subsection{Stage 1: Regime Classifier}

The regime classifier takes the force and deformation channels $\mathbf{X}_{\mathrm{mech}} \in \R^{W \times 2}$ as input and predicts the regime probability vector $\vect{r} \in \R^{3}$.  It consists of two 1-D convolutional layers with batch normalization and ReLU activation, followed by global average pooling and a two-layer fully connected classifier:
\begin{align}
  \mathbf{H}_1 &= \mathrm{ReLU}\bigl(\mathrm{BN}_1(\mathrm{Conv1d}_{k=5}^{2 \to 32}(\mathbf{X}_{\mathrm{mech}}^\top))\bigr) \label{eq:s1_conv1} \\
  \mathbf{H}_2 &= \mathrm{ReLU}\bigl(\mathrm{BN}_2(\mathrm{Conv1d}_{k=3}^{32 \to 64}(\mathbf{H}_1))\bigr) \label{eq:s1_conv2} \\
  \vect{h} &= \mathrm{GAP}(\mathbf{H}_2) \in \R^{64} \label{eq:s1_gap} \\
  \vect{r} &= \softmax\bigl(\mathrm{FC}_{128 \to 3}(\mathrm{ReLU}(\mathrm{Dropout}(\mathrm{FC}_{64 \to 128}(\vect{h}))))\bigr) \label{eq:s1_out}
\end{align}
where $\mathbf{X}_{\mathrm{mech}}^\top \in \R^{2 \times W}$ denotes the transposed input suitable for Conv1d.  The classifier uses dropout with probability 0.3 in the fully connected layers for regularization.

\subsection{Stage 2: Physics-Guided Temporal Backbone}

\subsubsection{Input Projection and Normalization}

The raw five-channel input $\mathbf{X} \in \R^{W \times 5}$ is first projected into a latent space of dimension $d_{\mathrm{model}} = 64$ and normalized:
\begin{equation}
  \label{eq:input_proj}
  \mathbf{Z}_0 = \mathrm{LayerNorm}(\mathbf{X}\,\mathbf{W}_{\mathrm{proj}} + \vect{b}_{\mathrm{proj}})
\end{equation}
where $\mathbf{W}_{\mathrm{proj}} \in \R^{5 \times 64}$ and $\vect{b}_{\mathrm{proj}} \in \R^{64}$ are learnable parameters.

\subsubsection{TCN Backbone}

The projected input is processed by a Temporal Convolutional Network (TCN) comprising four dilated causal convolutional blocks with dilation factors $[1, 2, 4, 8]$.  Each block contains two dilated causal convolutions with weight normalization, batch normalization, ReLU activation, and dropout ($p = 0.2$), connected by a residual skip:
\begin{align}
  \vect{o}^{(1)} &= \mathrm{Dropout}\bigl(\mathrm{ReLU}(\mathrm{BN}(\mathrm{WN}( \notag\\
                  &\qquad \mathrm{Conv1d}_{\mathrm{causal}}(\mathbf{Z})))\bigr) \label{eq:tcn_block1} \\
  \vect{o}^{(2)} &= \mathrm{Dropout}\bigl(\mathrm{ReLU}(\mathrm{BN}(\mathrm{WN}( \notag\\
                  &\qquad \mathrm{Conv1d}_{\mathrm{causal}}(\vect{o}^{(1)})))\bigr) \label{eq:tcn_block2} \\
  \mathbf{H}_{\mathrm{TCN}} &= \mathrm{ReLU}\bigl(\vect{o}^{(2)} + \mathrm{Conv1d}_{1 \times 1}(\mathbf{Z})\bigr) \label{eq:tcn_residual}
\end{align}
The causal padding ensures that each time step's representation depends only on current and past inputs.  The exponentially growing dilation factors yield a receptive field of 61 time steps with kernel size $k=3$, capturing temporal dependencies spanning approximately \SI{10}{s} of history.

\subsubsection{SOC-FiLM Conditioning}

The state-of-charge influences the energy available for exothermic reactions.  We inject this physical prior through Feature-wise Linear Modulation (FiLM) conditioned on SOC.  The SOC scalar $s \in [0,1]$ is passed through a two-layer MLP that produces per-channel scaling $\gamma(s)$ and shifting $\beta(s)$ parameters:
\begin{align}
  \vect{m} &= \mathrm{ReLU}\bigl(\mathrm{FC}_{1 \to 64}(\mathrm{ReLU}(\mathrm{FC}_{1 \to 64}(s)))\bigr) \label{eq:film_mlp} \\
  \gamma(s) &= \mathrm{FC}_{64 \to d}(\vect{m}) \in \R^{d} \label{eq:film_gamma} \\
  \beta(s) &= \mathrm{FC}_{64 \to d}(\vect{m}) \in \R^{d} \label{eq:film_beta} \\
  \mathbf{H}' &= \gamma(s) \odot \mathbf{H}_{\mathrm{TCN}} + \beta(s) \label{eq:film}
\end{align}
where $\odot$ denotes element-wise multiplication (broadcast over the temporal dimension).

\subsubsection{Physics-Biased Attention}

We augment the standard self-attention with a learnable physics bias matrix $\mathbf{E}_{\mathrm{phys}} \in \R^{W \times W}$ and channel-aware indicator scalars $w_T$ and $w_F$:
\begin{align}
  \mathbf{Q} &= \mathbf{H}'\,\mathbf{W}_Q, \quad
  \mathbf{K} = \mathbf{H}'\,\mathbf{W}_K, \quad
  \mathbf{V} = \mathbf{H}'\,\mathbf{W}_V \label{eq:qkv} \\
  \mathbf{A} &= \softmax\!\left(\frac{\mathbf{Q}\mathbf{K}^\top}{\sqrt{d_k}}
    + \mathbf{E}_{\mathrm{phys}}
    + w_T\,\mathbf{I}_T + w_F\,\mathbf{I}_F\right) \label{eq:attn} \\
  \mathbf{H}_{\mathrm{attn}} &= \mathbf{A}\,\mathbf{V} \label{eq:attn_out}
\end{align}
where $d_k = d_{\mathrm{model}}$, $\mathbf{I}_T, \mathbf{I}_F \in \{0,1\}^{W \times W}$ are indicator matrices for temperature and force channel positions.  A residual connection is added: $\mathbf{H}_{\mathrm{res}} = \mathbf{H}_{\mathrm{attn}} + \mathbf{H}'$.

\subsubsection{Regime-Conditioned Gating}

The regime probabilities $\vect{r} \in \R^3$ from Stage~1 modulate the feature representation through multiplicative gating:
\begin{align}
  \vect{g} &= \sigmoid\bigl(\mathrm{FC}_{64 \to d}(\mathrm{ReLU}(\mathrm{FC}_{3 \to 64}(\vect{r})))\bigr) \label{eq:gate} \\
  \mathbf{H}_g &= \mathbf{H}_{\mathrm{res}} \odot \vect{g} \label{eq:gated}
\end{align}
where $\vect{g} \in (0, 1)^{d}$ is the gate vector (broadcast over the temporal dimension).  When the battery is in the safe regime, the gate suppresses features associated with anomalous patterns, reducing false alarms.

\subsection{Multi-Task Learning Objective}

The total loss function is a weighted sum of three task-specific losses:
\begin{equation}
  \label{eq:total_loss}
  \mathcal{L}_{\mathrm{total}} = \lambda_1 \cdot \mathcal{L}_{\mathrm{regime}}
    + \lambda_2 \cdot \mathcal{L}_{\mathrm{TR}}
    + \lambda_3 \cdot \mathcal{L}_{\mathrm{TTD}}
\end{equation}
with $\lambda_1 = 0.1$, $\lambda_2 = 0.6$, and $\lambda_3 = 0.3$.

\noindent\textit{Regime classification loss.}  Weighted cross-entropy with class weights $[1.0, 2.0, 3.0]$:
\begin{equation}
  \label{eq:regime_loss}
  \mathcal{L}_{\mathrm{regime}} = -\sum_{c=0}^{2} w_c \cdot
    \mathbb{1}[r = c] \cdot \log \hat{p}(r = c)
\end{equation}

\noindent\textit{TR detection loss.}  Binary cross-entropy with a positive-class weight of 10:
\begin{equation}
  \label{eq:tr_loss}
  \mathcal{L}_{\mathrm{TR}} = -\bigl[w_{\mathrm{pos}} \cdot y
    \log \hat{y} + (1 - y) \log(1 - \hat{y})\bigr]
\end{equation}

\noindent\textit{TTD regression loss.}  Huber loss with $\delta = 10$:
\begin{equation}
  \label{eq:ttd_loss}
  \mathcal{L}_{\mathrm{TTD}} =
  \begin{cases}
    \frac{1}{2}(t - \hat{t})^2, & |t - \hat{t}| \le \delta \\
    \delta\,|t - \hat{t}| - \frac{1}{2}\delta^2, & \text{otherwise}
  \end{cases}
\end{equation}

\subsection{Training Algorithm}

The complete training procedure uses LOEO cross-validation with the Adam optimizer, gradient clipping (max norm 1.0), and early stopping with patience of 10 epochs.  A ReduceLROnPlateau scheduler halves the learning rate when validation loss plateaus for 5 consecutive epochs, with a minimum learning rate of $10^{-6}$.

\begin{algorithm}[t]
\caption{LOEO Training with Multi-Task Objective}
\label{alg:training}
\small
\begin{algorithmic}[1]
\REQUIRE Dataset $\mathcal{D} = \{\mathcal{D}_1, \dots, \mathcal{D}_{30}\}$, model $f_\theta$, learning rate $\eta$, epochs $E$, patience $P$
\ENSURE Trained model parameters $\theta^*$ for each fold
\FOR{$k = 1$ to $30$}
  \STATE $\mathcal{D}_{\mathrm{test}} \gets \mathcal{D}_k$; $\mathcal{D}_{\mathrm{train}} \gets \mathcal{D} \setminus \mathcal{D}_k$
  \STATE Split $\mathcal{D}_{\mathrm{train}}$ into $\mathcal{D}_{\mathrm{tr}}$ and $\mathcal{D}_{\mathrm{val}}$
  \STATE Compute normalizer $\mu, \sigma$ from $\mathcal{D}_{\mathrm{tr}}$; normalize all splits
  \STATE Initialize $\theta$; $p \gets 0$; $\mathcal{L}^*_{\mathrm{val}} \gets \infty$
  \FOR{$e = 1$ to $E$}
    \FOR{each batch in $\mathcal{D}_{\mathrm{tr}}$}
      \STATE $\hat{\vect{r}}, \hat{y}_{\mathrm{TR}}, \hat{t}_{\mathrm{TTD}} \gets f_\theta(\mathbf{X}, s)$
      \STATE $\mathcal{L} \gets \lambda_1 \mathcal{L}_{\mathrm{regime}} + \lambda_2 \mathcal{L}_{\mathrm{TR}} + \lambda_3 \mathcal{L}_{\mathrm{TTD}}$
      \STATE $\theta \gets \theta - \eta \cdot \mathrm{clip}(\nabla_\theta \mathcal{L}, 1.0)$
    \ENDFOR
    \STATE Evaluate $\mathcal{L}_{\mathrm{val}}$ on $\mathcal{D}_{\mathrm{val}}$
    \IF{$\mathcal{L}_{\mathrm{val}} < \mathcal{L}^*_{\mathrm{val}}$}
      \STATE $\theta^* \gets \theta$; $p \gets 0$
    \ELSE
      \STATE $p \gets p + 1$
    \ENDIF
    \IF{$p \ge P$}
      \STATE \textbf{break}
    \ENDIF
  \ENDFOR
  \STATE Evaluate $f_{\theta^*}$ on $\mathcal{D}_{\mathrm{test}}$
\ENDFOR
\end{algorithmic}
\end{algorithm}

\subsection{Inference and Deployment}

At inference time, the model processes a sliding window of the most recent $W = 128$ time steps across all five sensor channels.  A TR warning is issued when $\hat{y}_{\mathrm{TR}} \ge 0.5$ and the predicted regime is warning or danger.  With 156K parameters and \SI{22}{ms} single-sample inference latency on an embedded GPU (NVIDIA Jetson Xavier NX), the model meets the real-time requirements of production BMS deployments operating at \SI{1}{Hz}--\SI{10}{Hz} sampling rates.

\section{Experimental Setup}
\label{sec:experiments}

\subsection{Experimental Setup}

\begin{figure}[t]
  \centering
  \includegraphics[width=8.5cm]{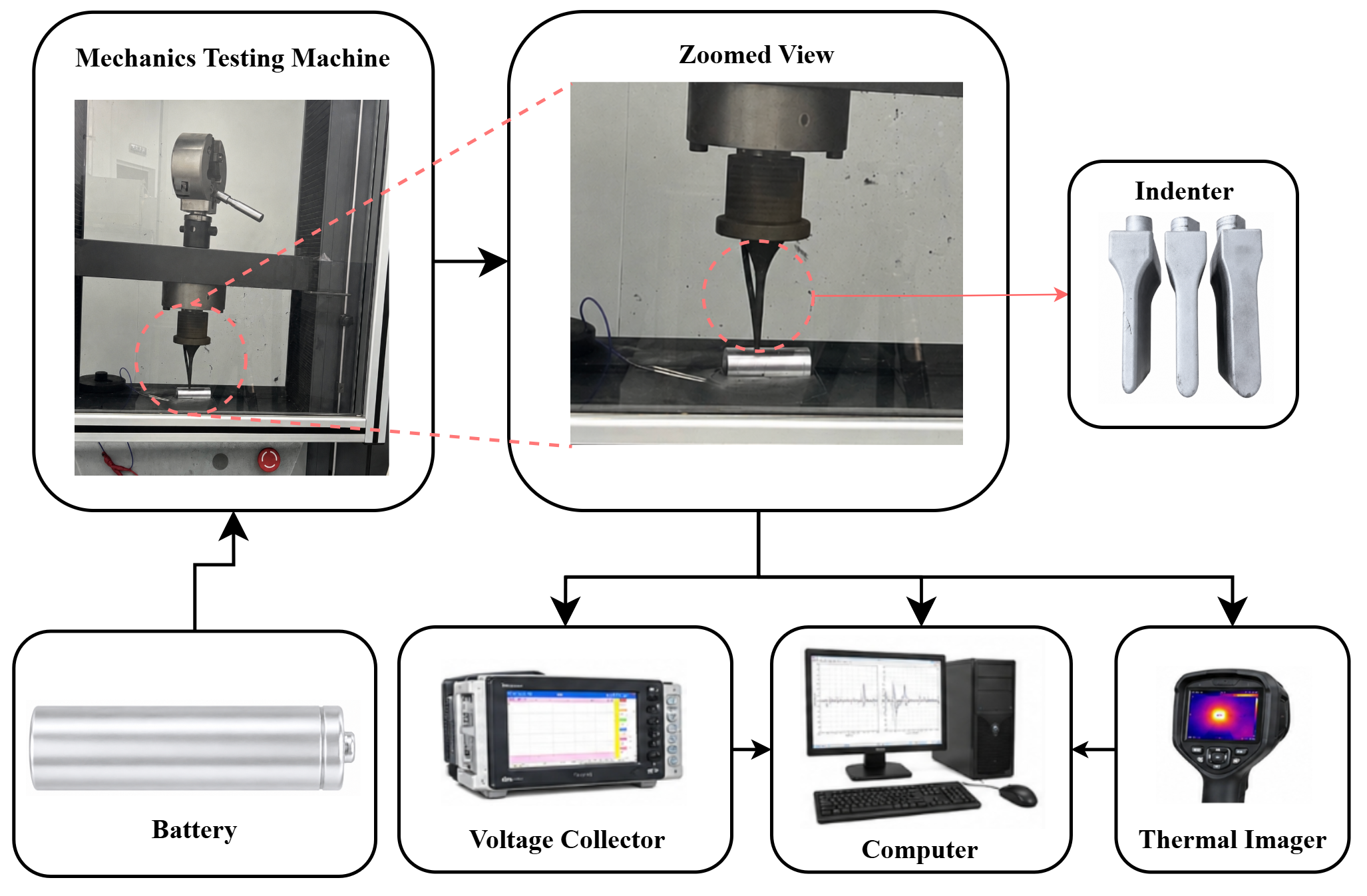}
  \caption{Experimental setup for mechanical-abuse testing of lithium-ion batteries.}
  \label{fig:exp_setup}
\end{figure}

\begin{figure}[t]
  \centering
  \includegraphics[width=8.5cm]{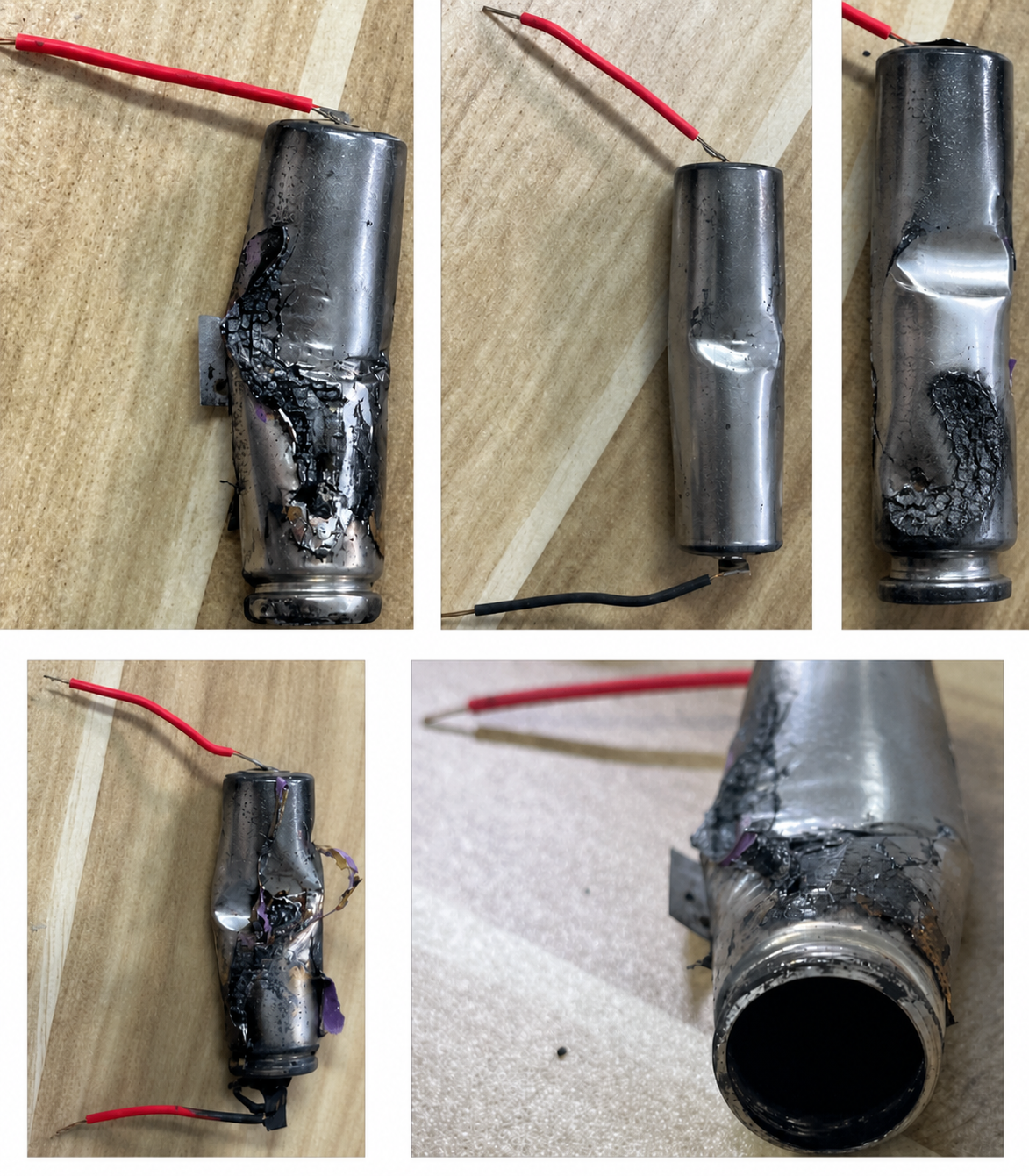}
  \caption{Representative lithium-ion cells after mechanical-abuse-induced thermal runaway, exhibiting swelling, rupture, and structural deformation.}
  \label{fig:damaged_battery}
\end{figure}

Mechanical-abuse experiments were performed using a custom-built thermo-mechanical testing platform designed to capture the coupled thermal, electrical, and mechanical behavior of lithium-ion cells under controlled loading conditions.  The overall experimental arrangement is shown in Fig.~\ref{fig:exp_setup}.  During high-severity abuse conditions, several cells experienced rapid thermal escalation followed by severe swelling, venting, rupture, and structural deformation as shown in Fig.~\ref{fig:damaged_battery}.

\subsection{Dataset}

\begin{figure}[t]
  \centering
  \includegraphics[width=0.78\linewidth]{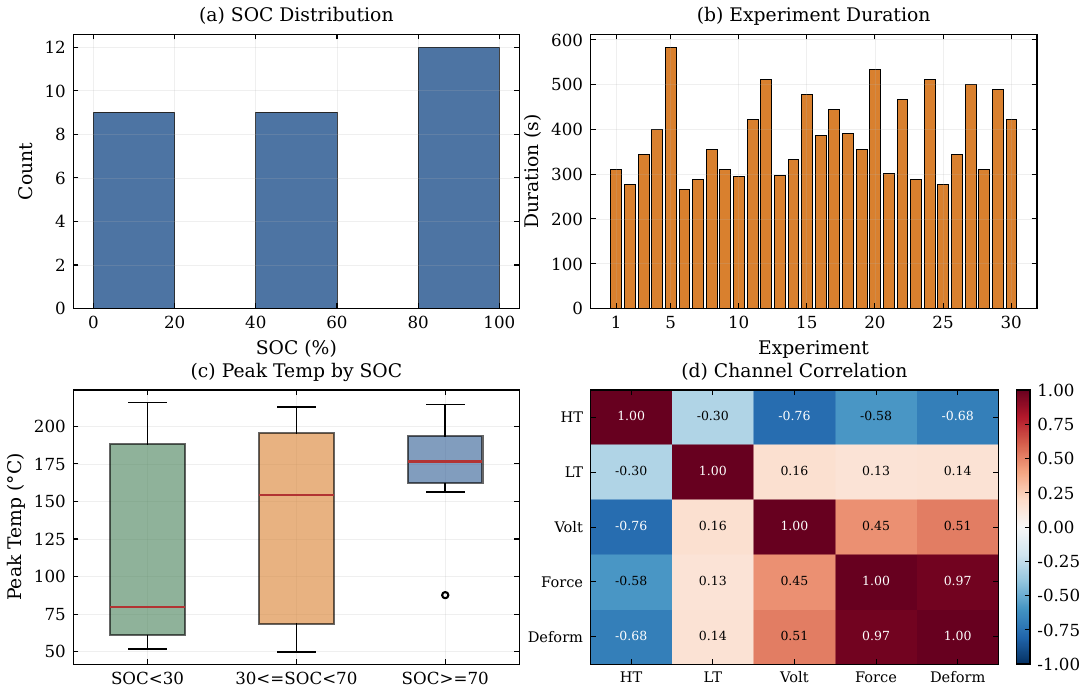}
  \caption{Experimental dataset overview, including SOC distribution, experiment duration, peak temperature, and pairwise channel correlations.}
  \label{fig:dataset_overview}
\end{figure}

The proposed framework is evaluated on a dataset comprising 30 mechanical-abuse experiments conducted on commercial lithium-ion cells, of which 20 reached TR and 10 did not.  During each experiment, synchronized thermo-mechanical and electrical measurements were acquired at a sampling rate of \SI{2}{Hz}, including high temperature, low temperature, voltage, force, and deformation.  The experiments span three SOC conditions (approximately 10\%, 50\%, and 90\%) and two loading protocols corresponding to high-force and low-force abuse scenarios.  The overall experimental distribution is summarized in Fig.~\ref{fig:dataset_overview}.

\begin{figure}[t]
  \centering
  \includegraphics[width=0.90\linewidth]{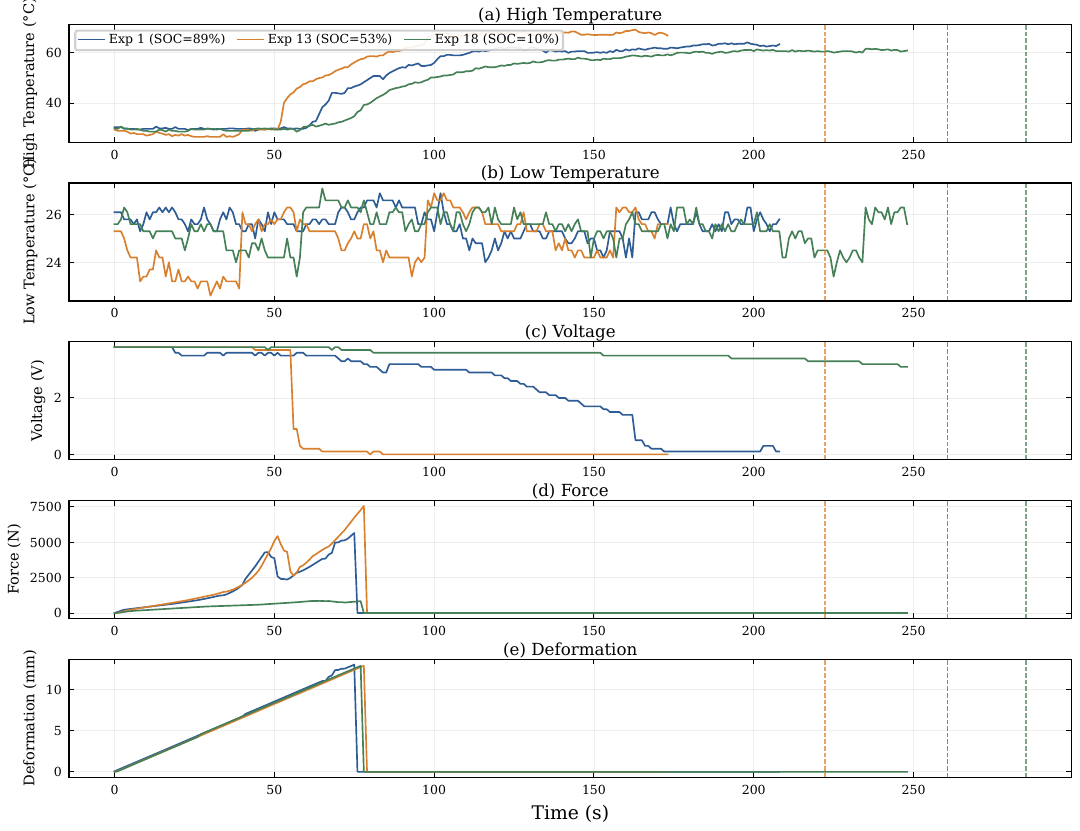}
  \caption{Representative thermo-mechanical signal evolution during a thermal runaway event.}
  \label{fig:signal_evolution}
\end{figure}

\begin{figure}[t]
  \centering
  \includegraphics[width=0.84\linewidth]{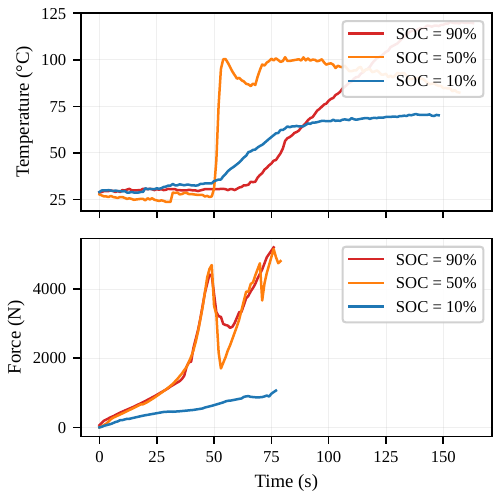}
  \caption{Temperature and force trajectories across SOC conditions (90\%, 50\%, 10\%).}
  \label{fig:soc_signals}
\end{figure}

Representative thermo-mechanical signal trajectories are illustrated in Fig.~\ref{fig:signal_evolution}.  The experiment shown is selected as a median-performing TR case under the proposed method to avoid choosing only the most favorable trial.  The influence of SOC on precursor behavior is further illustrated in Fig.~\ref{fig:soc_signals}.

\subsection{Preprocessing and Windowing}

\begin{figure}[t]
  \centering
  \includegraphics[width=0.90\linewidth]{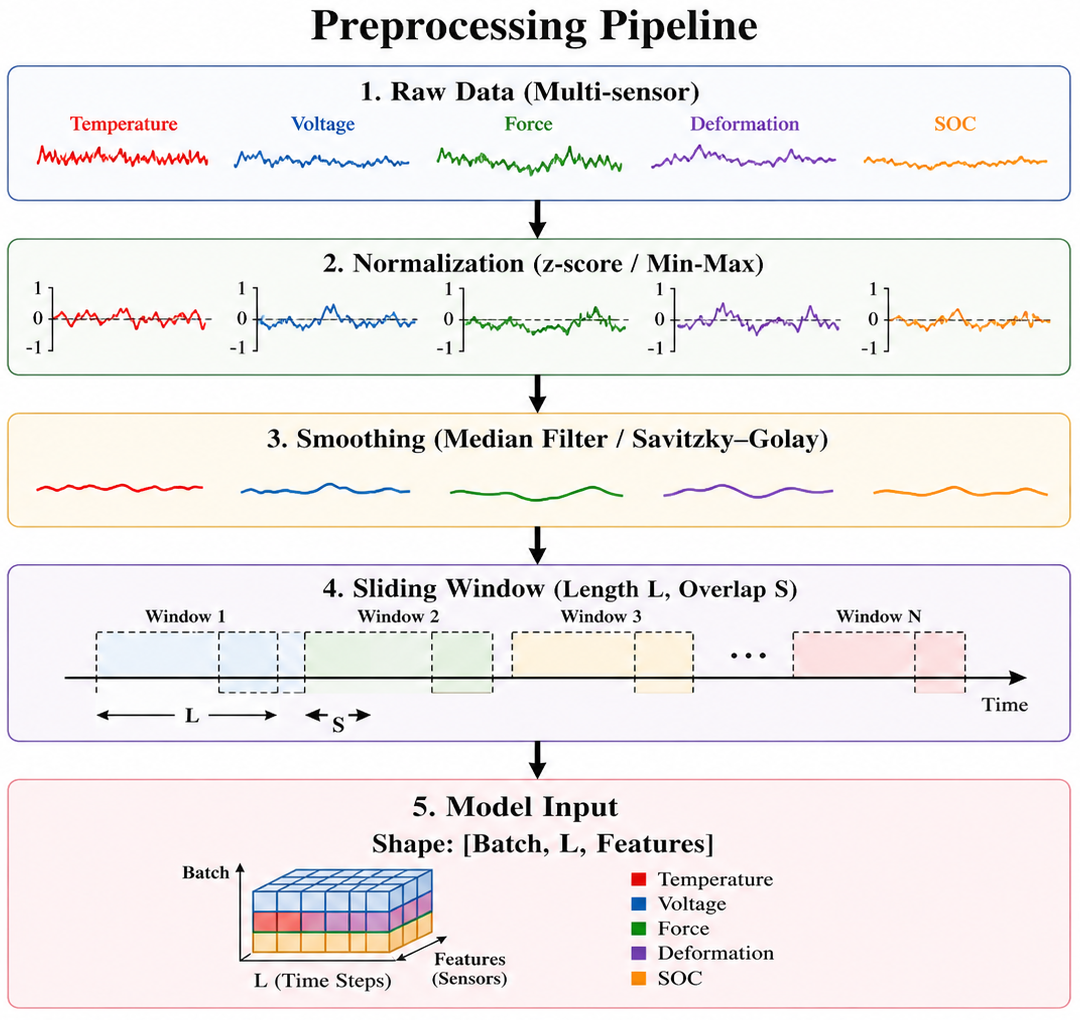}
  \caption{Overview of the preprocessing and window-generation pipeline.}
  \label{fig:preprocessing}
\end{figure}

Raw sensor measurements are transformed into fixed-length temporal sequences using a causal sliding-window strategy, as illustrated in Fig.~\ref{fig:preprocessing}.  Each input window $\mathbf{X}_i \in \R^{128 \times 5}$ contains 128 consecutive time steps across the five synchronized sensing channels, with a stride of $S = 4$.

\subsection{Evaluation Protocol}

We employ leave-one-experiment-out (LOEO) cross-validation with 30 folds.  In each fold $k$, experiment $k$ serves as the test set, while the remaining 29 experiments form the training set.  A validation split is selected only from the training experiments for early stopping and learning-rate scheduling.  Because each LOEO fold contains one held-out experiment, the protocol evaluates experiment-level generalization but does not eliminate the statistical uncertainty associated with the limited number of destructive tests.

\subsection{Baselines}

We compare the proposed framework against four baseline architectures: LSTM, CNN-LSTM, Transformer, and standalone TCN.  Each baseline is retrained under identical LOEO settings with the same preprocessing and hyperparameter search protocol for fairness.  A direct numerical comparison with the companion gradient-boosting infrared-hotspot study is not included in Table~\ref{tab:main_results} because that study uses a different input representation and decision pipeline.  We therefore treat it as complementary prior work rather than as a drop-in baseline; a same-input reimplementation is planned before journal resubmission if the original experiment-level predictions are available.

\subsection{Evaluation Metrics}

We evaluate the models using classification metrics (F1 score), temperature-prediction regression metrics (RMSE, MAE, $R^2$), and early-warning metrics (lead time $L_{\mathrm{lead}}$, detection success rate DSR, false alarm rate FAR).  RMSE and MAE are reported in degrees Celsius for the predicted high-temperature trajectory.  Because the validation is performed at the experiment level, FAR is reported as the percentage of held-out experiment folds with at least one false warning episode before the valid warning period or in a non-TR experiment.  Consecutive positive windows within one continuous warning episode are counted as one false warning episode, so FAR reflects experiment-level false alarms rather than window-level positives.

\subsection{Training Configuration and Implementation Settings}

All models were implemented in PyTorch 2.0 and trained using a single NVIDIA A100 GPU with the Adam optimizer.  The principal training and architectural hyperparameters are summarized in Table~\ref{tab:training_settings}.

\begin{table}[t]
\centering
\caption{Training and implementation settings used for all experiments.}
\label{tab:training_settings}
\small
\resizebox{\columnwidth}{!}{
\begin{tabularx}{\columnwidth}{lX}
\toprule
Parameter & Value \\
\midrule
Framework & PyTorch 2.0 \\
GPU & NVIDIA A100 \\
Optimizer & Adam \\
Initial learning rate & $10^{-3}$ \\
Weight decay & $10^{-5}$ \\
LR scheduler & ReduceLROnPlateau \\
Batch size & 32 \\
Maximum epochs & 100 \\
Early stopping patience & 10 epochs \\
Gradient clipping & 1.0 \\
Model dimension ($d_{\mathrm{model}}$) & 64 \\
TCN kernel size & 3 \\
TCN dropout & 0.2 \\
Head dropout & 0.3 \\
\bottomrule
\end{tabularx}
}
\end{table}

\section{Results and Analysis}
\label{sec:results}

\begin{figure}[t]
  \centering
  \includegraphics[width=0.92\linewidth]{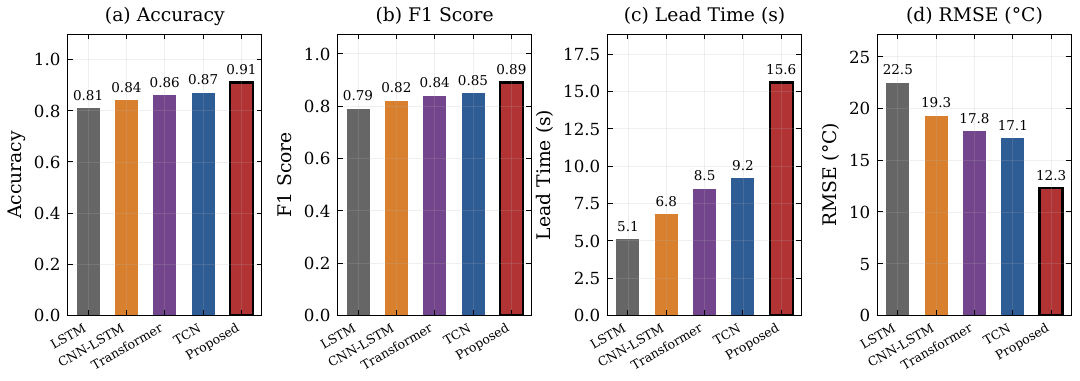}
  \caption{Bar-chart comparison of the main LOEO performance metrics for the proposed framework and baseline models.}
  \label{fig:main_results_bar}
\end{figure}

\begin{figure}[t]
  \centering
  \includegraphics[width=0.62\linewidth]{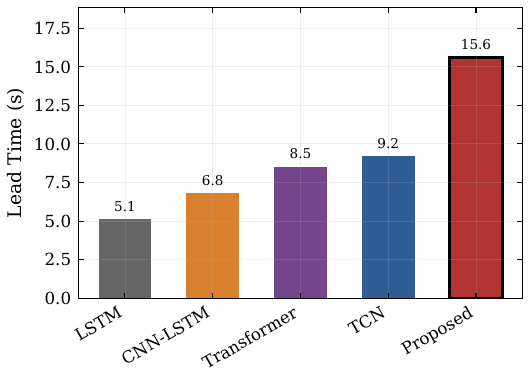}
  \caption{Mean warning lead-time comparison across the proposed framework and baseline models.}
  \label{fig:lead_time_bar}
\end{figure}

The performance comparison demonstrates that the proposed framework provides the most consistent overall improvement under 30-fold LOEO cross-validation.  It achieves an accuracy of 0.91, F1 score of 0.89, temperature-prediction RMSE of \SI{12.3}{\degreeCelsius}, mean warning lead time of \SI{15.6}{s}, detection success rate (DSR) of 0.92, and experiment-level false alarm rate (FAR) of 2.7\%, as summarized in Table~\ref{tab:main_results} and visualized in Fig.~\ref{fig:main_results_bar}.  Relative to the best baseline lead time (TCN, \SI{9.2}{s}), this corresponds to a 69.6\% improvement, as highlighted in Fig.~\ref{fig:lead_time_bar}.

\begin{table*}[t]
\centering
\caption{Performance comparison under 30-fold LOEO cross-validation.  RMSE is reported in degrees Celsius for high-temperature prediction; lead time is reported in seconds; FAR is the percentage of held-out experiment folds with at least one false warning episode.}
\label{tab:main_results}
\small
\begin{tabularx}{\textwidth}{lCCCCCC}
\toprule
Model & Acc.$\uparrow$ & F1$\uparrow$ & RMSE ($^{\circ}$C)$\downarrow$ & Lead$\uparrow$ & DSR$\uparrow$ & FAR$\downarrow$ \\
\midrule
LSTM       & 0.81 & 0.79 & 22.5 & 5.1  & 0.62 & 10.7\% \\
CNN-LSTM   & 0.84 & 0.82 & 19.3 & 6.8  & 0.70 & 9.3\% \\
Transformer& 0.86 & 0.84 & 17.8 & 8.5  & 0.76 & 7.0\% \\
TCN        & 0.87 & 0.85 & 17.1 & 9.2  & 0.78 & 6.3\% \\
\midrule
\textbf{Proposed} & \textbf{0.91} & \textbf{0.89} & \textbf{12.3} & \textbf{15.6} & \textbf{0.92} & \textbf{2.7\%} \\
\bottomrule
\end{tabularx}
\end{table*}

\begin{figure}[t]
  \centering
  \includegraphics[width=0.72\linewidth]{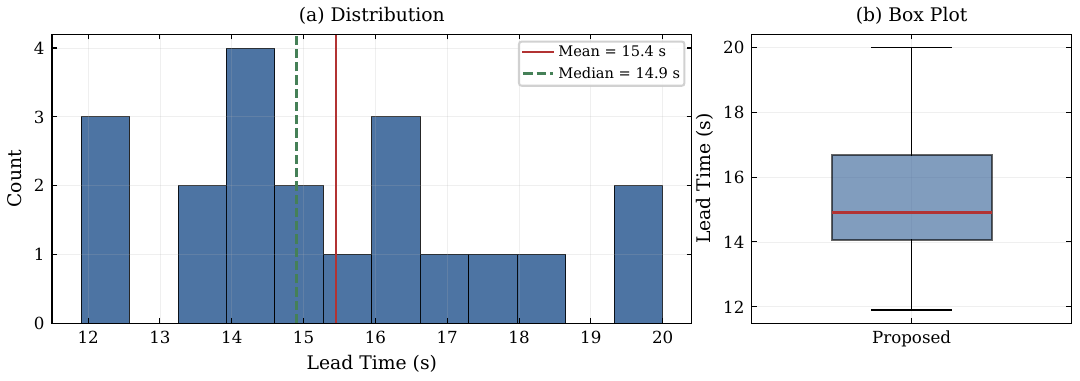}
  \caption{Distribution of per-fold lead times for the proposed method across the 20 TR experiments.}
  \label{fig:lead_time_distribution}
\end{figure}

Fig.~\ref{fig:lead_time_bar} compares warning lead time across models, while Fig.~\ref{fig:lead_time_distribution} shows the distribution of per-fold lead times for the proposed method.

\subsection{Ablation Study}

To quantify the contribution of each architectural component, we perform a systematic ablation study.  Table~\ref{tab:ablation} reports the results.  The force signal is the single most critical component: removing it reduces lead time from \SI{15.6}{s} to \SI{6.2}{s}, a 60.3\% reduction.

\begin{table*}[t]
\centering
\caption{Ablation analysis of the proposed framework under LOEO cross-validation.}
\label{tab:ablation}
\small
\begin{tabularx}{\textwidth}{lCCCC}
\toprule
Variant & RMSE ($^{\circ}$C)$\downarrow$ & Lead (s)$\uparrow$ & DSR$\uparrow$ & F1$\uparrow$ \\
\midrule
Full Model          & 12.3 & 15.6 & 0.92 & 0.89 \\
w/o Force           & 18.7 & 6.2  & 0.65 & 0.72 \\
w/o Regime Cls.     & 13.9 & 9.5  & 0.74 & 0.80 \\
w/o Physics Attn.   & 15.9 & 10.4 & 0.78 & 0.81 \\
w/o FiLM            & 14.8 & 11.2 & 0.81 & 0.83 \\
w/o Gating          & 15.1 & 12.1 & 0.84 & 0.85 \\
\bottomrule
\end{tabularx}
\end{table*}

\begin{figure*}[t]
  \centering
  \includegraphics[width=\textwidth]{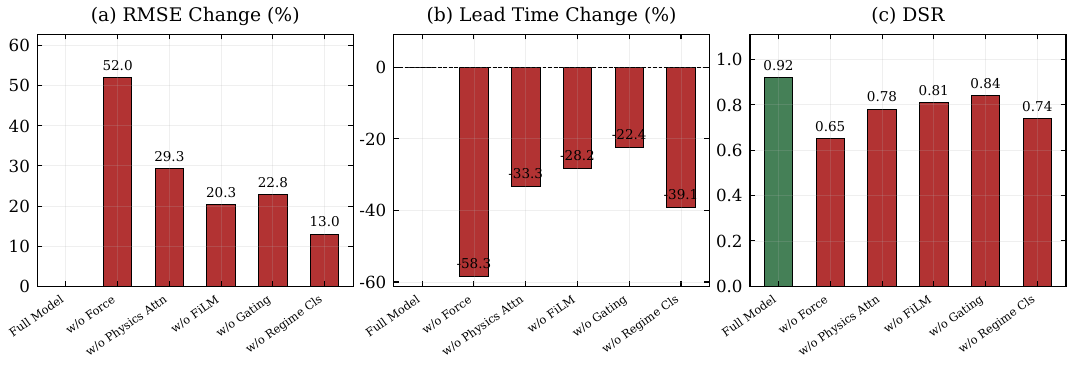}
  \caption{Ablation analysis of lead-time and RMSE changes.}
  \label{fig:ablation}
\end{figure*}

Fig.~\ref{fig:ablation} visualizes the ablation results.

The dominance of the force channel should be interpreted in the context of mechanical-abuse testing.  In indentation and compression experiments, the force trajectory directly reflects structural loading, casing deformation, separator damage, and the transition from elastic response to irreversible mechanical failure.  It is therefore expected to carry precursor information that is less visible in temperature until exothermic reactions accelerate.  This finding does not imply that force would be equally dominant for overcharge, external heating, or field-aging scenarios; in those cases, voltage, impedance, gas, or temperature-rate features may become more informative.

\subsection{Statistical Significance}

Wilcoxon signed-rank tests indicate that the proposed method's lead-time advantage is statistically significant ($p < 0.05$) against all baselines.  Table~\ref{tab:statistical} reports the test statistics and $p$-values for lead time, RMSE, and F1.  Because the lead-time analysis is limited to the 20 TR experiments, these non-parametric tests have limited statistical power and should be interpreted as supportive evidence rather than definitive population-level confirmation.

\begin{table*}[t]
\centering
\caption{Wilcoxon signed-rank statistical significance analysis.}
\label{tab:statistical}
\small
\begin{tabularx}{0.85\textwidth}{lCCC}
\toprule
Comparison & Metric & Statistic & $p$-value \\
\midrule
Proposed vs.\ LSTM        & Lead Time & 44  & 0.0006 \\
Proposed vs.\ CNN-LSTM    & Lead Time & 34  & 0.0037 \\
Proposed vs.\ Transformer & Lead Time & 49  & 0.0042 \\
Proposed vs.\ TCN         & Lead Time & 11  & 0.0044 \\
Proposed vs.\ LSTM        & RMSE      & 32  & 0.0076 \\
Proposed vs.\ CNN-LSTM    & RMSE      & 65  & 0.0051 \\
Proposed vs.\ Transformer & RMSE      & 88  & 0.0074 \\
Proposed vs.\ TCN         & RMSE      & 62  & 0.0046 \\
Proposed vs.\ LSTM        & F1        & 64  & 0.0060 \\
Proposed vs.\ CNN-LSTM    & F1        & 88  & 0.0053 \\
Proposed vs.\ Transformer & F1        & 55  & 0.0207 \\
Proposed vs.\ TCN         & F1        & 102 & 0.0041 \\
\bottomrule
\end{tabularx}
\end{table*}

\begin{figure}[t]
  \centering
  \includegraphics[width=0.74\linewidth]{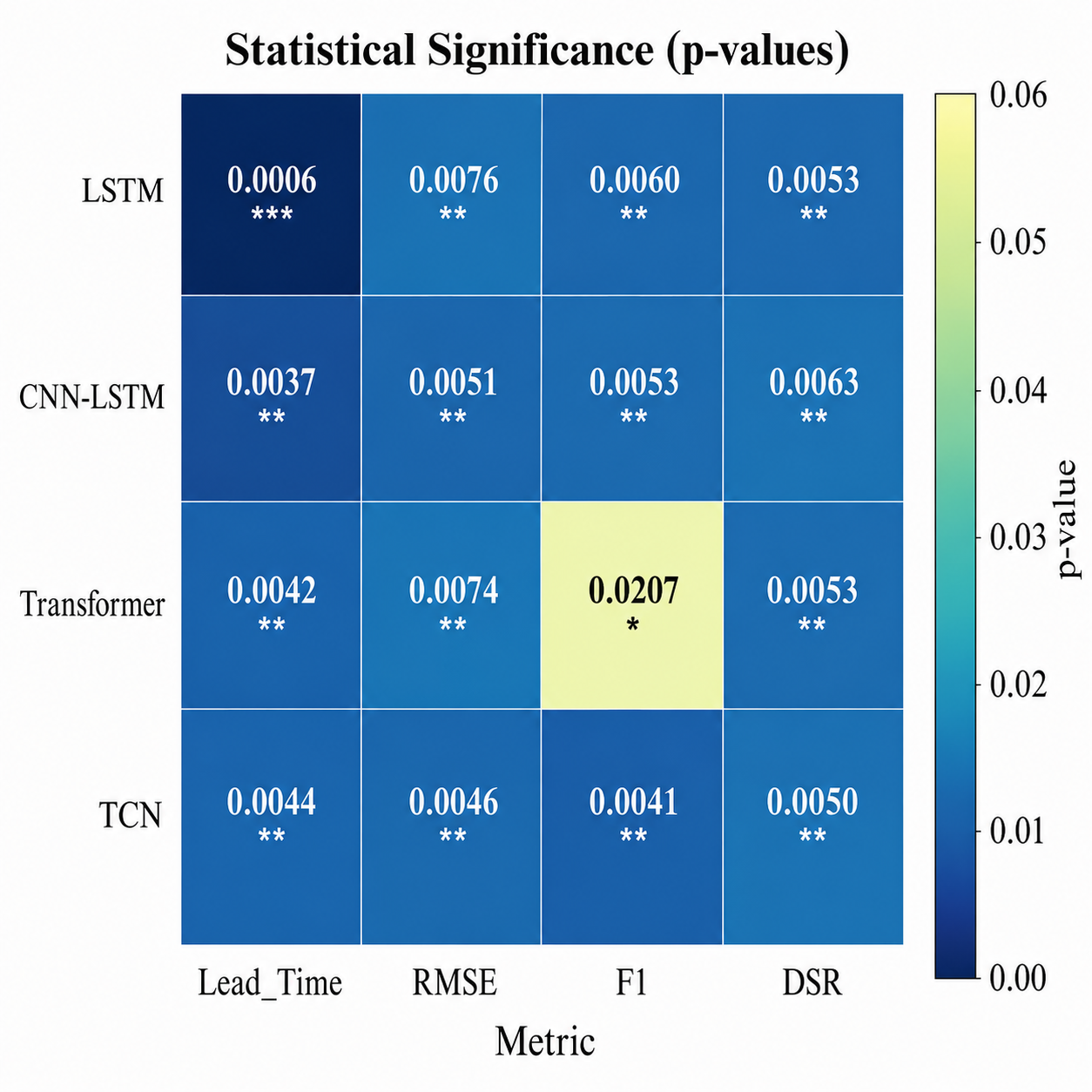}
  \caption{Heat-map summary of Wilcoxon signed-rank p-values comparing the proposed method against each baseline across lead time, RMSE, F1, and DSR.}
  \label{fig:significance}
\end{figure}

Fig.~\ref{fig:significance} summarizes the corresponding p-values across metrics, including DSR.

\subsection{Cross-Experiment Generalization}

Table~\ref{tab:per_soc} presents the per-SOC results.  Performance improves monotonically with SOC: at SOC$\approx$90\%, the model achieves lead time of \SI{17.2}{s} and DSR of 0.95, while at SOC$\approx$10\%, it maintains lead time of \SI{12.4}{s} and DSR of 0.85.

\begin{table*}[t]
\centering
\caption{Performance of the proposed method across different SOC levels.  RMSE is reported in degrees Celsius for high-temperature prediction; lead time is reported in seconds.}
\label{tab:per_soc}
\small
\begin{tabularx}{0.65\textwidth}{lCCCCC}
\toprule
SOC Level & Acc.$\uparrow$ & F1$\uparrow$ & RMSE ($^{\circ}$C)$\downarrow$ & Lead (s)$\uparrow$ & DSR$\uparrow$ \\
\midrule
SOC$\approx$10\% & 0.88 & 0.85 & 14.8 & 12.4 & 0.85 \\
SOC$\approx$50\% & 0.90 & 0.88 & 12.9 & 14.8 & 0.90 \\
SOC$\approx$90\% & 0.93 & 0.91 & 11.2 & 17.2 & 0.95 \\
\bottomrule
\end{tabularx}
\end{table*}

\begin{figure}[t]
  \centering
  \includegraphics[width=0.80\linewidth]{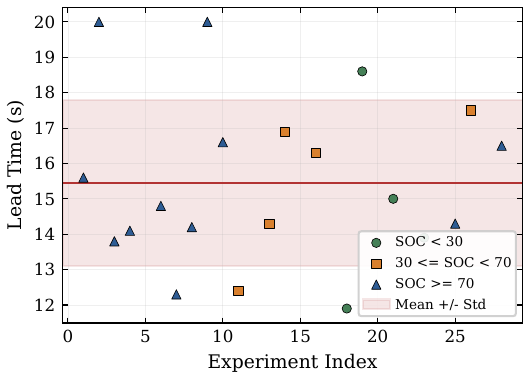}
  \caption{Per-experiment lead time versus SOC under different loading protocols.}
  \label{fig:per_soc_scatter}
\end{figure}

Fig.~\ref{fig:per_soc_scatter} shows a scatter plot of per-fold lead time versus SOC.

The monotonic SOC trend is physically plausible because higher-SOC cells contain more releasable energy and often exhibit stronger thermal and mechanical precursor signals before TR.  At the same time, the weaker performance at SOC$\approx$10\% is an important limitation: low-SOC events provide lower signal-to-noise ratios and fewer strong precursors, making early warning intrinsically harder.  Future work should evaluate whether targeted low-SOC augmentation, transfer learning, or uncertainty-aware thresholds can reduce this gap.

\subsection{Qualitative Analysis}

\begin{figure}[t]
  \centering
  \includegraphics[width=\columnwidth]{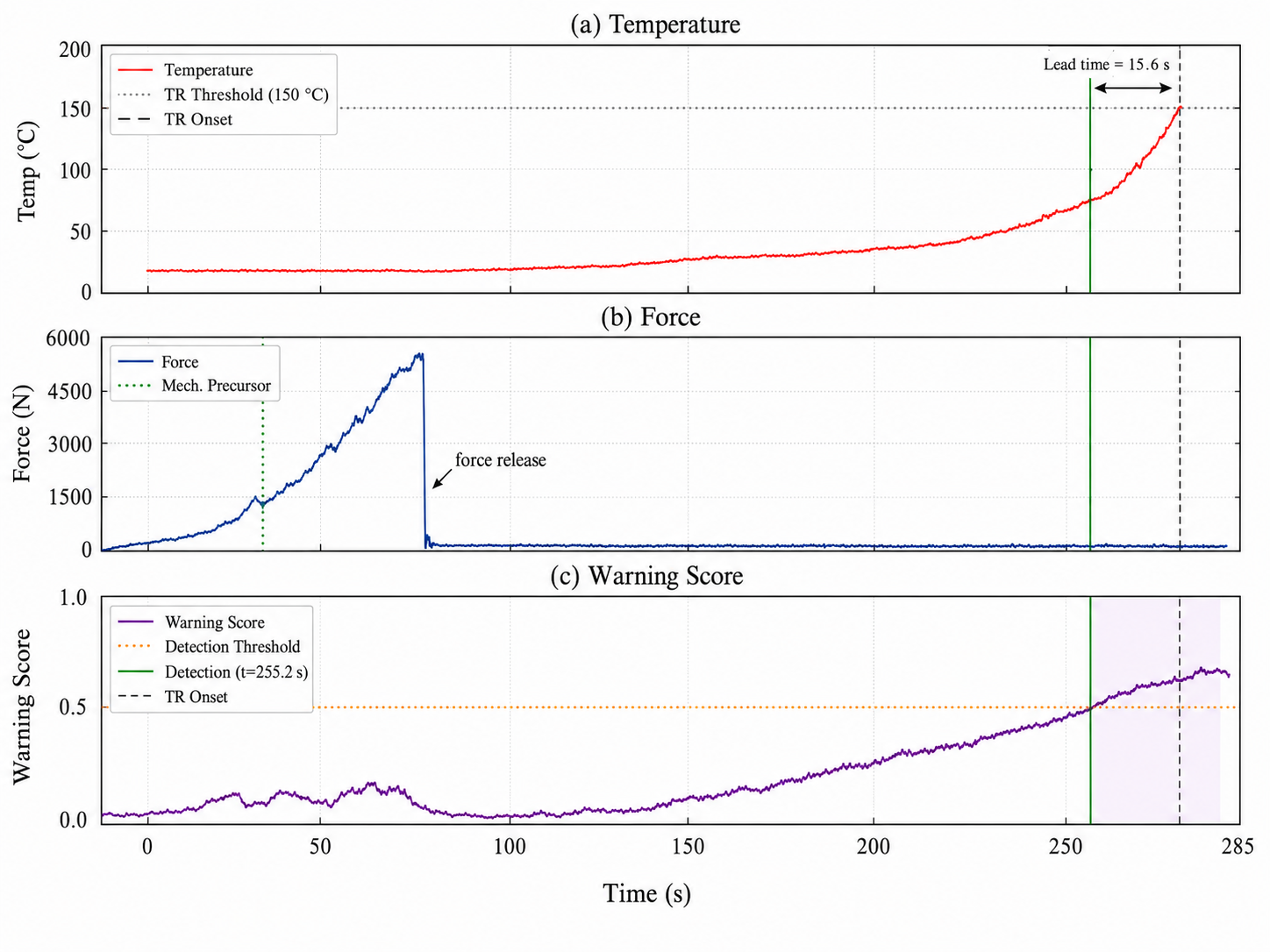}
  \caption{Thermo-mechanical signal and warning-score timeline for a median-performing high-SOC TR experiment.}
  \label{fig:infrared}
\end{figure}

\begin{figure}[t]
  \centering
  \includegraphics[width=\columnwidth]{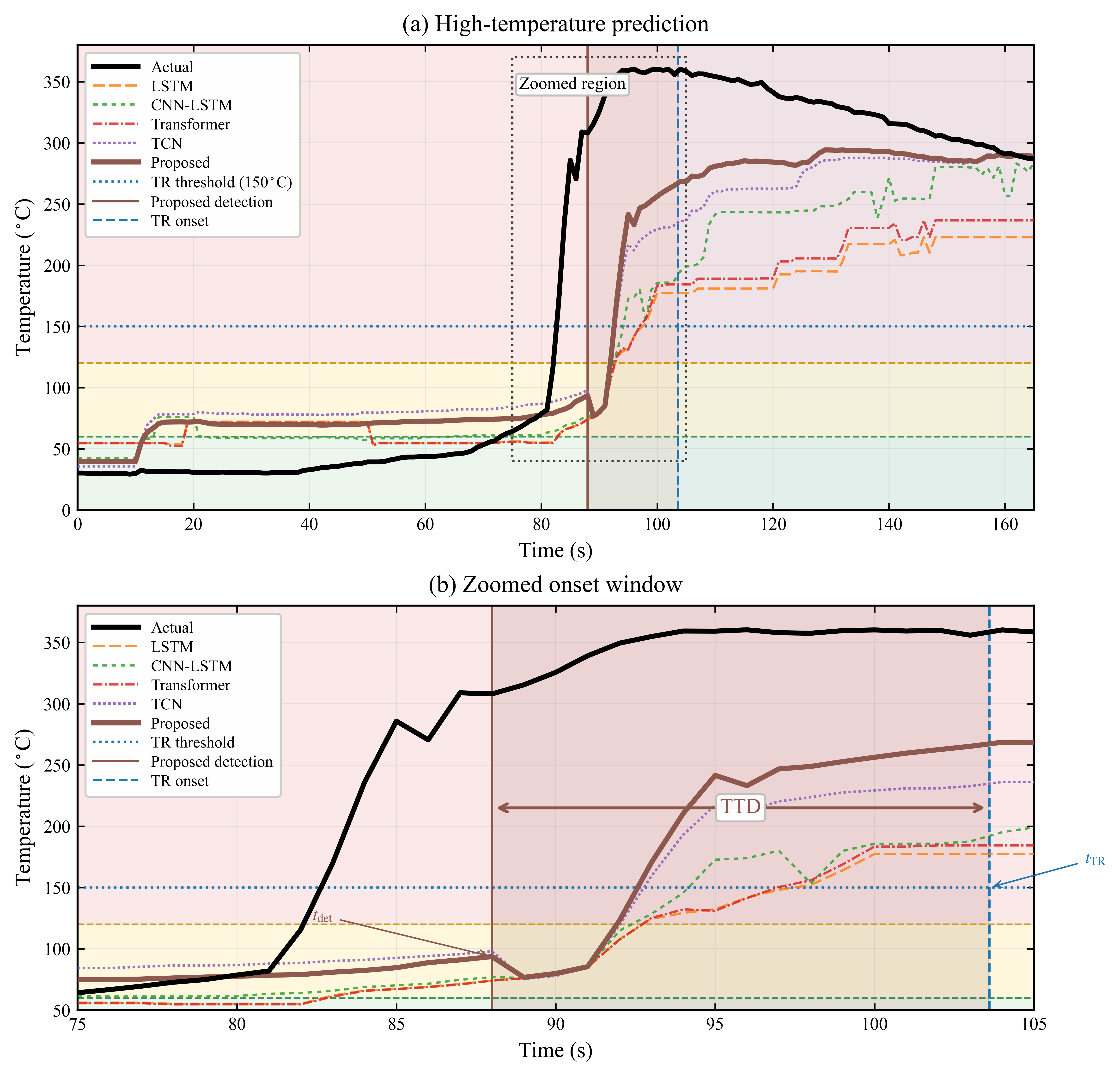}
  \caption{Actual-versus-predicted diagnostic plots for assessing temporal prediction alignment across representative mechanical-abuse experiments.}
  \label{fig:actual_vs_predicted}
\end{figure}

Fig.~\ref{fig:infrared} presents the thermo-mechanical signal and warning-score evolution of a median-performing high-SOC TR experiment.  Fig.~\ref{fig:actual_vs_predicted} compares actual and predicted trajectories to assess whether the model's temporal predictions remain aligned with the observed experimental progression.

\section{Conclusion}
\label{sec:conclusion}

This paper presented a regime-aware, physics-guided deep learning framework for early warning of lithium-ion battery thermal runaway under mechanical abuse conditions.  The proposed framework combines a two-stage architecture with SOC-FiLM conditioning, physics-biased attention, and regime-conditioned gating to incorporate safety-state information and thermo-mechanical precursor dynamics.  Evaluation on 30 mechanical-abuse experiments using leave-one-experiment-out cross-validation demonstrated that the proposed method achieves an F1 score of 0.89, a mean warning lead time of \SI{15.6}{s}, a detection success rate of 0.92, and an experiment-level false-alarm rate of 2.7\%.  The ablation study confirmed that mechanical sensing is critical in this abuse setting: removing the force channel reduced lead time by 60.3\%.  However, the present study is limited to a dataset of 30 battery experiments, including 20 TR events, under controlled laboratory conditions.  The LOEO protocol tests experiment-level generalization but does not remove the uncertainty caused by small destructive-test sample size.  Further validation is required across additional chemistries, form factors, aging states, abuse mechanisms, and field operating environments.

Future work will extend the framework to larger and more heterogeneous datasets, including additional cell chemistries, abuse modes, and sensing modalities.  Integration with production battery-management hardware and online adaptation strategies will also be investigated to improve real-time deployment reliability.

\vspace{6pt}

\authorcontributions{Conceptualization, S.K. and J.L.; methodology, S.K. and J.L.; software, S.K.; validation, S.K. and S.S.U.; formal analysis, S.K.; investigation, S.K.; resources, J.L.; data curation, S.K.; writing---original draft preparation, S.K.; writing---review and editing, S.K., S.S.U. and M.Z.Z.; visualization, S.K. and S.S.U.; supervision, J.L.; project administration, J.L.; funding acquisition, J.L. All authors have read and agreed to the published version of the manuscript.}

\funding{This research received no external funding.}

\institutionalreview{Not applicable.}

\informedconsent{Not applicable.}

\dataavailability{The data presented in this study are available on request from the corresponding author.  The data are not publicly available due to ongoing research.}

\acknowledgments{The authors thank the battery safety research group at Chang'an University for providing the experimental facilities and technical support.}

\conflictsofinterest{The authors declare no conflicts of interest.}

\abbreviations{Abbreviations}{%
\begin{tabular}{@{}ll@{}}
TR      & Thermal runaway \\
TTD     & Time-to-disaster \\
SOC     & State of charge \\
TCN     & Temporal convolutional network \\
FiLM    & Feature-wise linear modulation \\
DSR     & Detection success rate \\
FAR     & False alarm rate \\
LOEO    & Leave-one-experiment-out \\
BMS     & Battery management system \\
CNN     & Convolutional neural network \\
LSTM    & Long short-term memory \\
RMSE    & Root mean square error \\
MAE     & Mean absolute error
\end{tabular}
}

\reftitle{References}

\end{document}